\pdfoutput=1
\documentclass[11pt]{article}
\usepackage{ACL2023}
\usepackage{times}
\usepackage{latexsym}
\usepackage[T1]{fontenc}
\usepackage[utf8]{inputenc}
\usepackage{microtype}
\usepackage{inconsolata}
\usepackage{bm}
\usepackage{xcolor}
\usepackage{amsfonts}
\usepackage{todonotes}
\usepackage{natbib}
\usepackage{graphicx}
\usepackage{footnote}
\usepackage{color}
\usepackage{amsfonts}
\usepackage{multicol}
\usepackage{multirow}
\usepackage{amssymb}
\usepackage{booktabs}
\usepackage{enumitem}
\usepackage{physics}
\usepackage{makecell}
\usepackage{stfloats}

\title{DiffuDetox: A Mixed Diffusion Model for Text Detoxification}

\author{Griffin Floto \and Mohammad Mahdi Abdollah Pour \\ \and Parsa Farinneya \and Zhenwei Tang \and Scott Sanner \\
  University of Toronto \\
  \texttt{\{griffin.floto, zhenwei.tang,m.abdollahpour,parsa.farinneya\}@mail.utoronto.ca} \\ \AND
  Ali Pesaranghader \and Manasa Bharadwaj \\
  LG Electronics, Toronto AI Lab \\
  \texttt{\{ali.pesaranghader, manasa.bharadwaj\}@lge.com} \\}

\author{
    Griffin Floto\textsuperscript{1}, Mohammad Mahdi Abdollah Pour\textsuperscript{1}, Parsa Farinneya\textsuperscript{1}, Zhenwei Tang\textsuperscript{1}, \\ {\bf Ali Pesaranghader\textsuperscript{2}, Manasa Bharadwaj\textsuperscript{2},} \and {\bf Scott Sanner\textsuperscript{1}} \\
    \textsuperscript{1} University of Toronto, Canada \\
    \resizebox{\linewidth}{!}{
        \texttt{\{griffin.floto,m.abdollahpour,parsa.farinneya,zhenwei.tang\}@mail.utoronto.ca}
    } \\
    \texttt{ssanner@mie.utoronto.ca} \\
    \textsuperscript{2} LG Electronics, Toronto AI Lab \\
    \texttt{\{ali.pesaranghader, manasa.bharadwaj\}@lge.com}
}

\begin{document}
\maketitle

\begin{abstract}
Text detoxification is a conditional text generation task aiming to remove offensive content from toxic text. It is highly useful for online forums and social media, where offensive content is frequently encountered.
Intuitively, there are diverse ways to detoxify sentences while preserving their meanings, and we can select from detoxified sentences before displaying text to users.
Conditional diffusion models are particularly suitable for this task given their demonstrated higher generative diversity than existing conditional text generation models based on language models.
Nonetheless, text fluency declines when they are trained with insufficient data, which is the case for this task.
In this work, we propose DiffuDetox\footnote{\url{https://github.com/D3Mlab/diffu-detox}}, 
a mixed \textit{conditional} and \textit{unconditional} diffusion model for text detoxification. 
The conditional model takes toxic text as the condition and reduces its toxicity, yielding a diverse set of detoxified sentences. The unconditional model is trained to recover the input text, which allows the introduction of additional fluent text for training and thus ensures text fluency.
Extensive experimental results and in-depth analysis demonstrate the effectiveness of our proposed DiffuDetox.

\end{abstract}

\section{Introduction}

Toxic texts with offensive and abusive words are frequently encountered in online forums and social media. Such a harmful online environment can lead to mental health problems \cite{viner2019roles, wijesiriwardene2020alone}, which motivates considerable research efforts \cite{dos2018fighting, laugier2021civil, logacheva2022paradetox} in text detoxification, i.e., a conditional text generation task aiming to remove offensive content from sentences while preserving their meanings. 

Intuitively, there exist diverse ways to detoxify a given sentence.
As shown in Table~\ref{fig:example},
some detoxified sentences are the results of simply removing or replacing the toxic word, e.g., Detoxified 1 and 2, which may cause loss of information or lower text fluency.
While other candidates, e.g., Detoxified 3, can reach human-level text detoxification performance with satisfactory fluency and content preservation.
Therefore, if a diverse collection of detoxified sentences are given, we can select the most fluent and preservative one to maximize user experience.
To do so, we resort to textual conditional diffusion models \cite{difflm, gong2022diffuseq} because they are shown to be capable of generating more diverse sets of candidates compared to existing solutions based on transformers \cite{vaswani2017attention}, e.g., GPT2 \cite{radford2019language}.
Given their demonstrated high generative diversity, diffusion models are particularly suitable for this task.

\begin{table}[t!]
\small 
  \centering
    \begin{tabular}{m{1.6cm}|m{5cm}}
    \Xhline{0.225ex}
    \textbf{Toxic} & The country doesn't really have to \textbf{give a shit} about international laws.\\
    \hline \hline
    \textbf{Detoxified 1} & The country doesn't really have to give \textbf{[$\cdots$]} about international laws. \\ \hline
    \textbf{Detoxified 2} & The country doesn't really have \textbf{care} about international laws. \\ \hline
    \textbf{Detoxified 3} & The country doesn't really \textbf{need to care} about international laws.\\ 
    \hline \hline
    \textbf{Human} &  The country doesn't need to care about international laws. \\
    \Xhline{0.225ex}
    \end{tabular}%
    \caption{A diverse collection of detoxified sentences helps to approach human-level text detoxification.}
\label{fig:example}
\end{table}

Nevertheless, previous textual conditional diffusion models \cite{difflm, gong2022diffuseq} are not directly applicable to text detoxification due to the scarcity of text detoxification data. Given that text detoxification is a relatively new field and the high cost of human annotations, the available text detoxification data is on the order of $1e^{-1}$ to $1e^{-2}$ of datasets used for other tasks with textual conditional diffusion models \cite{gong2022diffuseq}. 

To this end, we introduce DiffuDetox, a mixed \textit{conditional} and \textit{unconditional} diffusion model for text detoxification. 
In particular, the conditional model takes toxic text as a condition and through a Markov chain of diffusion steps, yields a diverse set of detoxified sentences.
On the other hand, the unconditional model is trained to recover any given input text exactly. That allows us to introduce additional fluent text to be reconstructed by the unconditional model, which is used to improve the fluency of the conditionally generated detoxified sentences.
In this way, the resulting diffusion model can maintain a diverse collection of detoxified candidates with satisfactory sentence fluency and content preservation. 
Extensive experimental results and in-depth discussions demonstrate the effectiveness of DiffuDetox for text detoxification. 
Our main contributions are summarized in two folds: 1) To the best of our knowledge, we are the first to approach text detoxification with diffusion models, which can maintain a rich collection of detoxified sentences by their high generative diversity; 2) We propose a mixed diffusion model for text detoxification, where the conditional model reduces text toxicity and the unconditional model improves text fluency.



\begin{figure*}[t!]
    \centering
    \resizebox{\linewidth}{!}{
    \includegraphics[trim={0.75cm 0 0 0},clip]{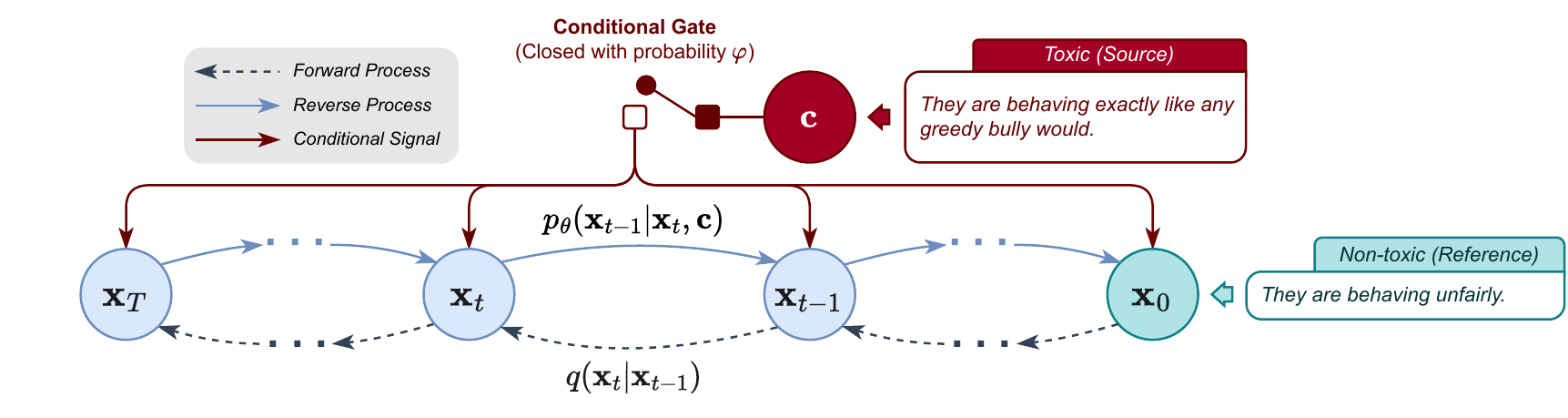}
    }
    \caption{The overall framework of DiffuDetox, a mixed conditional and unconditional diffusion model. For the conditional learning phase, the condition gate is closed with probability $\varphi$, then $\vb{x}_0$ and $\vb{c}$ are sampled from the detoxification dataset. $\vb{x}_0$ and $\vb{c}$ are set as non-toxic text and toxic text, respectively. For the unconditional learning phase, the condition gate is open with probability $1 - \varphi$, and $\vb{x}_0$ is sampled from the fluent text corpus.
    }
    \label{diffudetox-diagram}
\end{figure*}


\section{Related Work}

\subsection{Text Detoxification}
\label{baselines}
Previous text detoxification efforts fall into two main categories, \textit{supervised} and \textit{unsupervised}.
The unsupervised methods are built on a set of toxic and a set of non-toxic texts without one-to-one mappings between them. Representative methods include Mask\&Infill \cite{wu2019mask}, DRG-Template/Retrieve \cite{li2018delete}, DLSM \cite{he2020probabilistic}, SST \cite{lee2020stable}, CondBERT and ParaGeDi \cite{dale2021text}.
In contrast, the supervised methods are built on parallel datasets in which one-to-one mappings between toxic and non-toxic texts are explicitly provided. ParaDetox \cite{logacheva2022paradetox} is a well-established method within this category, which fine-tunes BART \cite{lewis2020bart} on their parallel data.



\subsection{Textual Diffusion Models}

Diffusion probabilistic models are deep generative models with Markov chains of diffusion steps to recover the noise slowly added to data \cite{sohl}.
Recently, diffusion models have shown impressive performance on \textit{continuous} domains such as image and audio generation \cite{ddpm, wave}, sparking interest in using these models in \textit{discrete} spaces like text. 
Some textual diffusion models use a discrete diffusion process that operates on word tokens \cite{step, diffuser}, whereas other methods convert text to embeddings, and then treat text as continuous variables \cite{difflm, will}. 
Although textual diffusion models have proved to be effective in various text generation tasks with rich data \cite{gong2022diffuseq}, they have not yet been applied to tasks with fewer training samples, such as text detoxification in our case. \citet{cfg} are the first to exploit unconditional diffusion models for conditional generation, while their method is limited to images and is not aiming for introducing additional data under the low-data setting.

\section{Methodology}
As the overall framework of DiffuDetox shown in Figure \ref{diffudetox-diagram} details, our proposed diffusion model for text detoxification improves text fluency in the low-training data regime by using a mixture of a conditional and unconditional diffusion model. We overview diffusion models before discussing DiffuDetox in detail.

\subsection{Diffusion Models}
Diffusion is a generative modeling paradigm that can be understood as a denoising algorithm \cite{sohl, song, score_based}. 
Noise is gradually added to data samples, while the diffusion model is trained to reverse the process and recover the original data. The framework can be described as a Markov process with $T$ steps, where the original data exist at $t=0$. Given a sample $\vb{x}_0$, the so-called forward process gradually adds noise to the data points, i.e., the blue arrows in Figure \ref{diffudetox-diagram}. The noisy sample can be described by: 
\begin{equation}
    q(\vb{x}_t | \vb{x}_{t-1}) := \mathcal{N}(\vb{x}_{t}; \sqrt{1-\beta_t}\vb{x_t}, \beta_t\vb{I})
\end{equation}
where the variance schedule parameters $\beta_1,\cdots,\beta_T$ are selected such that $\beta_t \in [0,1]$ and $\beta_0$ is close to $0$ and $\beta_T$ is close to $1$ \cite{ddpm}. This ensures that when $t \approx 0$, the data has little noise added to it, while when $t\approx T$, the data is identical to a sample from a standard Gaussian distribution.

The reverse process then attempts to remove the noise that was added in the forward process and is parameterized by $\theta$ as: 
\begin{equation}
    p_{\theta}(\vb{x}_{t-1} | \vb{x}_t) := \mathcal{N}(\vb{x}_{t-1}; \vb{\mu}_{\theta}(\vb{x}_t,t), \sigma_t\vb{I})
\end{equation}
where the predictive model $\vb{\mu}_{\theta}$ is: 
\begin{equation}
\label{equ:pred-model}
    \vb{\mu}_{\theta} := \frac{1}{\sqrt{\alpha_t}}(\vb{x}_t - \frac{\beta_t}{\sqrt{1 - \bar\alpha_t}} \vb{\epsilon}_{\theta}(\vb{x}_t, t))
\end{equation}
which depends on time-dependent coefficients $\alpha := 1 - \beta_t$, $\bar{\alpha}_t := \prod_{s=1}^t \alpha_s$. In Eq.\ \eqref{equ:pred-model}, $\vb{\epsilon}_{\theta}$ is interpreted as predicting the noise that was added to $\vb{x}_t$. To optimize the log-likelihood of this model, a simplified training objective is used which reduces the problem to: 
\begin{equation}
\small
    \mathcal{L} = \mathbb{E}_{t,\vb{x}_0, \vb{\epsilon}} [\Vert \vb{\epsilon} - \vb{\epsilon}_{\theta}(\sqrt{\bar{\alpha}_t} \vb{x}_0 + \sqrt{1 - \bar{\alpha}_t} \vb{\epsilon}, t) \Vert^2]
\end{equation}
After training, samples are generated by beginning with pure noise from a standard Gaussian distribution, which is then gradually denoised $T$ times by the learned reverse process.

\subsection{DiffuDetox: A Mixed Diffusion Model for Text Detoxification}
\label{diffudetox}

The task of text detoxification can be viewed as generating a non-toxic sentence, conditioned on a toxic input sentence. The goal is to ensure that the semantics and content of the text are preserved after detoxification, while ensuring that the generated text is fluent. With this interpretation \cite{gong2022diffuseq}, we can apply a conditional diffusion model that generated non-toxic text, when conditioned on a toxic sentence. A conditional diffusion model is modified such that the reverse process is now $p_{\theta}(\vb{x}_{t-1} | \vb{x}_t, \vb{c})$, and the predictive model is $\vb{\epsilon}_{\theta}(\vb{x}_t, \vb{c}, t)$. This model can be interpreted as mapping sequences to sequences in a non-autoregressive manner. To apply this model to textual data, sentences are tokenized and converted to a stack of embeddings which are then taken to be $\vb{x}_0$ in the diffusion process. When sampling, embeddings that are generated by the diffusion model are converted to tokens by a shallow single-layer decoder. 

While diffusion models have high sample diversity which can be used to generate a large number of candidate items, the fluency of the samples is degraded when trained on a smaller dataset. We propose to use a combination of the conditional model diffusion model as well as an unconditional model to tackle this problem. The conditional model is used to detoxify text, whereas the unconditional model can be used to guide the sampling process towards higher quality samples \cite{cfg}. The models are combined in a manner that is inspired by the gradient of an implicit classifier $p^i(\vb{c}|\vb{x}) \propto p(\vb{x}|\vb{c}) / p(\vb{x})$ such that the following linear combination of the models is used for sampling: 
\begin{equation}
\label{equ:diffudetox}
    \bar{\vb{\epsilon}}_{\theta}(\vb{x},\vb{c}) = (1+w) \vb{\epsilon}_{\theta}(\vb{x},\vb{c}) - w \vb{\epsilon}_{\theta}(\vb{x})
\end{equation}

\section{Experiments}
\label{sec:experiments}





\subsection{Experimental Settings}
\paragraph{Datasets.}
We conduct our experiments upon a well-established benchmarking dataset {ParaDetox}\footnote{\url{https://huggingface.co/datasets/SkolkovoInstitute/paradetox}} \cite{logacheva2022paradetox}, which provides human-annotated one-to-one mappings of toxic and non-toxic sentence pairs from 20,437 paraphrases of 12,610 toxic sentences. We use the same data split of \citet{logacheva2022paradetox} with 671 testing sentences for fair performance comparisons.
We further consider the BookCorpus \cite{Zhu_2015_ICCV}, MNLI \cite{wang2019glue}, and WikiAuto \cite{acl/JiangMLZX20}, datasets as additional data for unconditional diffusion model training. 


\paragraph{Evaluation Metrics.}
We follow the well-established text detoxification work \cite{logacheva2022paradetox} to evaluate DiffuDetox with BLEU, Style Accuracy (STA), Content Preservation (SIM), Fluency (FL), and J score.
In particular, STA and FL are computed with pre-trained classifiers \cite{warstadt2019neural} to measure the non-toxicity and fluency of a given sentence, respectively. And we compute SIM using cosine similarity between the input and the generated detoxified text with the model of \citet{wieting2019beyond}. Moreover, we compute J score \cite{krishna2020reformulating} as the averaged multiplication of STA, SIM, and FL, which is highly correlated with human evaluation as shown by \citet{logacheva2022paradetox}.

\paragraph{Implementation Details. }
We implement our mixed conditional and unconditional models with a single diffusion model where $c=\emptyset$ for the unconditional case. During training, the conditional model is selected with probability $\varphi=0.8$, and the unconditional model is trained using the non-toxic sentences sampled from the ParaDetox dataset and the additional dataset with equal probabilities. We use the union of the BookCorpus, WikiAuto, and MNLI as the additional dataset. In the test stage, we select the best samples from a candidate set of $20$ using the J score. The reported results are from a model trained for $1e^5$ steps with a batch size of 32, and the mixture weighting parameter $w$ in Eq.\ \eqref{equ:diffudetox} is set to $5$. We use the text detoxification methods listed in Section \ref{baselines} as baselines.

\subsection{Experimental Results}
\paragraph{Performance Comparison.}
\label{sec:exp-res}
We have two key observations from the results shown in Table \ref{tab:comp-study}. 
Firstly, our proposed DiffuDetox outperforms most baseline methods on most evaluation metrics, and it is reaching state-of-the-art performance by outperforming ParaDetox on two metrics, demonstrating the effectiveness of our proposed method.
Another observation is that DiffuDetox achieves a higher J score than human-level text detoxification. Note that the J score has been shown to be highly correlated with human annotations \cite{logacheva2022paradetox}. This human-level performance of DiffuDetox shows its promise to be deployed in real-world text detoxification scenarios to facilitate users in online forums and social media. Moreover, such results are achieved by selecting from the diverse collection of detoxified sentences generated by diffusion models, which reveals their high generative diversity and the suitability of being applied to text detoxification.
Examples of detoxified sentences generated by DiffuDetox can be found in Appendix \ref{sec:appendix}.

\paragraph{Ablation Study.}
\label{sec:comp-study}
We conduct ablations study to investigate the effectiveness of the unconditional model. Since the unconditional model allows the introduction of the additional fluent text, the ablation study can provide insights into the effect of both the unconditional model and the introduced additional data. As shown in Table \ref{tab:comp-study}, the model named \textit{Conditional} represents DiffuDetox without the unconditional component. 
We observe that the addition of the unconditional model improves all the metrics. In particular, text fluency achieves the most significant performance gain. More importantly, the addition of the unconditional model pushes the diffusion model over the human baseline for the J score. Such results demonstrate the effectiveness of the unconditional model and the introduced additional fluent text in improving text fluency and overall performance.

\begin{table}[t!]
\small
  \centering

  \resizebox{\columnwidth}{!}{%
  \renewcommand{\arraystretch}{1.1}
    \begin{tabular}{l||c|c|c|c||c}
    \Xhline{0.225ex}
          & \textbf{BLEU} & \textbf{STA } & \textbf{SIM } & \textbf{FL } & \textbf{J} \\
    \hline
    Human & 100.0 & 0.96 & 0.77 & 0.88 & 0.66 \\
    \hline
    DRG-Template & 53.86 & 0.90  & 0.82  & 0.69  & 0.51 \\
    DRG-Retrieve & 4.74  & 0.97  & 0.36  & 0.86  & 0.31 \\
    Mask\&Infill & 52.47 & 0.91  & 0.82  & 0.63  & 0.48 \\
    CondBERT & 42.45 & 0.98  & 0.77  & 0.88  & 0.62 \\
    SST   & 30.20 & 0.86  & 0.57  & 0.19  & 0.10 \\
    ParaGeDi & 25.39 & \underline{\textbf{0.99}}  & 0.71  & 0.88  & 0.62 \\
    DLSM  & 21.13 & 0.76  & 0.76  & 0.52  & 0.25 \\
    ParaDetox & \underline{\textbf{64.53}} & 0.89  & \underline{0.86}  & \underline{\textbf{0.89}}  & \underline{\textbf{\textit{0.68}}} \\
    \hline \hline
    Conditional & 61.43 & 0.91 & 0.87 & 0.78 & 0.64 \\
    \hline
    \textbf{DiffuDetox} & 62.13 & 0.92 & \textbf{0.88} & 0.80 & \textit{0.67} \\
    \Xhline{0.225ex}
    \end{tabular}%
    }
\caption{Text detoxification performance on the ParaDetox dataset. Baseline results are taken from \cite{logacheva2022paradetox}. The best results are in boldface, the strongest baseline performance is underlined, and the J score results reaching human-level detoxification performance are in italics.}
\label{tab:comp-study}
\end{table}%



\section{Conclusion}
In this paper, we approach the text detoxification task with diffusion models for their demonstrated high generative diversity.
We introduced DiffuDetox, a mixed conditional and unconditional diffusion model, where the conditional part reduces toxicity whereas the unconditional part ensures fluency. 
Experimental results show DiffuDetox achieves human-level text detoxification performance, making it promising to be applied in real-world text detoxification systems to benefit users.


\section*{Limitations and Future Work}
One limitation of our method is that sampling requires sampling both a conditional and a unconditional model, which results in slower inference times. On the other hand, progressive distillation \cite{distill} provides an attractive solution to this problem. Another limitation is that \citet{cfg} show that the diversity of generative models is degraded as $w$ increases. Ideally we would be able to have a model that improves upon the fluency as well as the model diversity.

As for future work, we will leverage advanced large language models as the base architecture for training diffusion models to compete with high performance auto-regressive models. Additionally, we will investigate modifications to diffusion models that are inherent to discrete data.

\section*{Ethics Statement}
\paragraph{Potential Misuse:} DiffuDetox can hypothetically be used to obtain toxic sentences from non-toxic sentences. However, the effectiveness of such a scenario should be investigated. 
\paragraph{Environmental Cost:} We note that while our work
required extensive experiments to draw sound conclusions, future work will be able to draw on these insights and need not run as many large-scale comparisons. Models in production may be trained once using the most promising settings.


\section*{Acknowledgements}
We would like to acknowledge that this work was supported by LG Electronics, Toronto AI Lab Grant Ref No. 2022-1473.

\bibliography{anthology,custom}
\bibliographystyle{acl_natbib}

\newpage

\appendix

\section{Appendix}
\label{sec:appendix}

Table \ref{tab:examples} shows examples of toxic texts with DiffuDetox paraphrases and human references. DiffuDetox is able to achieve human-level paraphrasing performance as evaluated quantitively in Section \ref{sec:exp-res}.
\begin{table*}[]

\resizebox{\linewidth}{!}{
\begin{tabular}{|l|l|}
\hline
                & Example                                                                                 \\ \hline
Toxic Sentence          & that guy used to be a real dick.    \\
DiffuDetox      &  that guy used to be a real bad man.   \\
Human Reference &  That guy used to be a mean person.        \\\hline
Toxic Sentence          &  i didn t say shit!\\
DiffuDetox      &  i didn t say anything! \\
Human Reference & I did not say that.\\\hline
Toxic Sentence           & one isn t dead but just totally fucked up.                                             \\
DiffuDetox      & one isn t dead but just totally messed up.                                              \\
Human Reference & One is not dead but just completely destroyed.                                           \\\hline
Toxic Sentence          & kid, you should be watching the crap they call cartoons now made for your generation . \\
DiffuDetox      & kid, you should be watching the stuff they call cartoons now made for your generation.  \\
Human Reference & kid, you should be watching the cartoons now made for your generation.                \\\hline
Toxic Sentence          & spoiler russia doesn t give a fuck what canada thinks.                                 \\
DiffuDetox      & spoiler russia doesn t care what canada thinks.                                         \\
Human Reference & Russia doesn't care what Canada thinks.                                                 \\\hline
Toxic Sentence          & how the fuck do you support the economy when you can barely support yourself.          \\
DiffuDetox      & how do you support the economy when you can barely support yourself.                    \\
Human Reference & How to support the economy when you can barely support yourself                         \\\hline
Toxic Sentence          & harper needs to get the fuck out of office.                                            \\
DiffuDetox      & harper needs to get out of office.                                                      \\
Human Reference & Harper needs to get out of office                                                      \\\hline
Toxic Sentence          &    again , give me the name of the store or fuck off, liar.\\
DiffuDetox      &  again, give me the name of the store or go away.  \\
Human Reference &  again, give me the name of the store.\\\hline

Toxic Sentence          & now that is just a fucking dumb thing to say. \\
DiffuDetox      &   now that is just a bad thing to say. \\
Human Reference & now that is just a useless thing to say.   \\\hline

\end{tabular}
}
\caption{Examples for performance comparison of DiffuDetox against human reference}
\label{tab:examples}
\end{table*}

\end{document}